\documentclass[conference]{IEEEtran}
\IEEEoverridecommandlockouts
\usepackage{subcaption}
\usepackage{xurl}
\usepackage[utf8]{inputenc}
\usepackage[T1]{fontenc}
\usepackage{lmodern}
\usepackage{microtype}
\usepackage[scaled]{helvet}
\usepackage[english, vietnamese]{babel}
\usepackage{graphicx}
\usepackage[bookmarks=false]{hyperref}
\usepackage{comment}
\usepackage{vietnam}
\usepackage{fancyhdr}
\usepackage{float}
\usepackage{cite}
\usepackage{mathptmx}
\usepackage{amsmath,amssymb,amsfonts}
\usepackage{algorithmic}
\usepackage{graphicx}
\usepackage{textcomp}
\usepackage{xcolor}
\usepackage{times} 
\usepackage{setspace}
\usepackage{graphicx}
\usepackage{caption}
\def\BibTeX{{\rm B\kern-.05em{\sc i\kern-.025em b}\kern-.08em
    T\kern-.1667em\lower.7ex\hbox{E}\kern-.125emX}}

\addto\captionsvietnam{\renewcommand{\abstractname}{Abstract}}
\addto\captionsvietnam{\renewcommand{\refname}{References}}
\begin{document}

\immediate\write18{for i in *.ps; do ps2pdf -DEPSCrop \$i; done;}

\title{A Big Data-empowered System for Real-time Detection of Regional Discriminatory Comments on Vietnamese Social Media}

\author{
  \IEEEauthorblockN{An-Nghiep Huynh\textsuperscript{*,\textdagger}, 
  Thanh-Dat Do\textsuperscript{*,\textdagger},
  Trong-Hop Do \textsuperscript{*,\textdagger} }
  \IEEEauthorblockA{\textsuperscript{*} University of Information Technology \\ \textsuperscript{\textdagger}Vietnam National University Ho Chi Minh City,Vietnam}
  \{21522377, 21520694\}@gm.uit.edu.vn, \\ Corresponding author: hopdt@uit.edu.vn
}\maketitle

%%==================================%%
%% sample for unstructured abstract %%
%%==================================%%

\begin{abstract}
% Discriminatory and prejudiced comments on social media platforms can perpetuate negative stereotypes. In the context of Vietnam, regional discrimination is a prevalent issue that needs to be addressed. This project aims to detect and classify regional discriminatory comments on social media, utilizing data crawling techniques on social media, machine learning, and deep learning models. The research aims to create an effective classification model and proposes a system that can automatically detect and visualize regional discriminatory comments on social media platforms. The study also suggests potential actions for platform management and offers valuable insights to promote a more positive and inclusive online environment.
Regional discrimination is a persistent social issue in Vietnam. While existing research has explored hate speech in the Vietnamese language, the specific issue of regional discrimination remains under-addressed. Previous studies primarily focused on model development without considering practical system implementation. In this work, we propose a task called Detection of Regional Discriminatory Comments on Vietnamese Social Media, leveraging the power of machine learning and transfer learning models. We have built the ViRDC (Vietnamese Regional Discrimination Comments) dataset, which contains comments from social media platforms, providing a valuable resource for further research and development. Our approach integrates streaming capabilities to process real-time data from social media networks, ensuring the system's scalability and responsiveness. We developed the system on the Apache Spark framework to efficiently handle increasing data inputs during streaming. Our system offers a comprehensive solution for the real-time detection of regional discrimination in Vietnam.

\end{abstract}
\section{Introduction} 
\subsection{Background}
% Social media has transformed how Vietnamese people connect and express themselves, bringing both benefits and challenges. While it fosters communication, it also amplifies issues like regional discrimination and prejudice.

% Vietnam's rich cultural diversity can sometimes lead to online tensions and misunderstandings. Discriminatory comments perpetuate stereotypes and create social divides, impacting user experiences and contributing to a toxic digital environment. Vulnerable groups, especially from specific regions, may face harassment and marginalization.

The state of discrimination and prejudice in society often relates to specific characteristics such as race, gender, health status, occupation, education, and living standards. Regional discrimination is defined as the prejudice or bias of one social group against another based on their place of origin or hometown. The issue of regional discrimination exists to varying degrees in most countries around the world. After 1945, the problem of discrimination, prejudice, and bias became evident in countries such as the United States, Canada, some European countries, and Australia \cite{bertrand2004emily}. In Germany, discrimination between citizens of the East and West persists more than 25 years after reunification. In Vietnam, after several prolonged wars that divided the country over the past centuries \cite{wikipedia}, regional discrimination has increased. This threatens national unity and development, causes psychological harm to victims, and provides opportunities for hostile forces to undermine the current political system of the country.   

Social media platforms, while allowing people to connect and communicate, have also become places where discriminatory behavior spreads easily.\cite{vtv:phan_biet_vung_mien} On December 8, 2023, VTV24, a news program produced by the Center for Digital Content Production and Development, Vietnam Television, aired a special episode addressing the issue of regional discrimination on social media platforms in Vietnam. Emphasizing its impact on national unity. 

 There are several state-of-the-art approaches for text classification in the Vietnamese language, including machine learning, and deep learning. However, existing research mainly focuses on developing models without implementing practical systems, which are crucial for use in social networks. To address this gap, we propose a system that leverages big data technology to continuously collect and process data. Our models are deployed on the Apache Spark framework to ensure the system can handle the increasing volume of online data. Additionally, we introduce ViRDC, a dataset specifically created for the automatic detection of regional discrimination in Vietnamese social media comments, serving as a crucial asset for future research and development. 
 
The paper's structure is as follows: We shall discuss pertinent research on the issue in Section 2. In the following section, we will give a thorough overview of ViRDC, covering the methods for collecting data, categorizing them, and organizing the dataset. The model used in the study will be explained in Section 4. We will go into detail on the model's architecture, training algorithms, and selective methods. In Section 5, the experiments carried out on the dataset will be covered in detail, along with the results and analytical assessments that follow. After that, we will explain the real-time prediction features and the streaming data procedure in Section 6. Section 7 will conclude the study by providing a summary of the results and a proposal for future research.
\subsection{Motivation}
Research on detecting regional discriminatory comments on social media in Vietnam, using Spark and Kafka, is motivated by critical goals. The importance of detecting regional discriminatory comments in Vietnam lies not only in the social aspect but also has profound scientific and practical significance. In the context of the growing development of social networks, regional discriminatory comments not only cause harm to individuals but also increase division within the community. Research and detection of these comments help raise awareness of the issue, while also providing crucial data to develop automated tools to identify and prevent negative content. Furthermore, this research contributes to building a healthy online environment, promoting solidarity and mutual understanding among regions in Vietnam.

% \subsection{Tool, Datasets used in paper}
% The research incorporates a set of robust tools and datasets to conduct a comprehensive investigation. The primary development environment, Google Colab, serves as a collaborative and cloud-based platform for executing various research tasks. Selenium, a powerful web testing framework, is employed for web scraping and data extraction, enabling automated interaction with web pages. Additionally, Kafka, a distributed streaming platform, has been seamlessly integrated into the research workflow. 

% A pivotal element of this study is the recently curated dataset named ViRDC (Vietnamese Regional Discrimination Comments). This dataset, purposefully gathered for the research, specifically focuses on comments related to regional discrimination within the Vietnamese context.
% \subsection{Structure of the remaining paper}

\section{Related Work}
The study on “Hate Speech Detection on Vietnamese” \cite{luu2021large}, \cite{quoc2023vietnamese} offers crucial insights for our research. The meticulous two-phase data preprocessing, including lowercase conversion, link removal, and de-teen code mapping, provides a solid foundation for handling social media comments. The success of the PhoBERT-CNN model demonstrates the efficacy of combining transfer learning and deep learning.

These findings prompt a reevaluation of our preprocessing strategies and model architectures. The F1 scores of 64.43\% and 90.89\% on ViHSD and HSD-VLSP datasets serve as benchmarks for our hate speech detection models. As we progress, the lessons from this study will enhance our approach to addressing similar challenges.

\section{Dataset}
\subsection{Data collection}
\subsubsection{Source}
The ViRDC (Vietnamese Regional Discrimination Comments) dataset is a meticulously self-collected compilation of around 17,000 social media comments. Sourced from posts and videos, ViRDC aims to research and analyze language in social media, focusing on regional discrimination in Vietnam. This dataset offers unparalleled flexibility and control, ensuring accuracy and representativeness in studying the Vietnamese social media language, especially regarding regional issues. The link to the dataset can be found here: \href{https://github.com/anhuynh219/bigdata}{https://github.com/anhuynh219/bigdata}
 
ViRDC is categorized into three label types to represent the diversity of social media language, Tables \ref{table example data_V} with Vietnamese language and \ref{table: example data_E} illustrate the three types of data in the dataset.
\renewcommand{\tablename}{Table}
\begin{table}[h!]
\centering
\caption{The labels in dataset}
\centering
\renewcommand{\arraystretch}{1.5}
\begin{tabular}{clc}
    \hline
   \# \textbf{Label} & \textbf{Describe} & \textbf{Example}\\
    \hline
    0.0 & Other & haha this is so funny\\
    1.0 & Discrimination & All Northerners are assh*le \\
    2.0 & Supportive & North and South are one family.\\
    \hline
\end{tabular}
\label{table: example data_E}
\end{table}

\renewcommand{\tablename}{Table}
\begin{table}[h!]
\centering
\caption{The labels in the dataset are presented in Vietnamese}
  \renewcommand{\arraystretch}{1.5}
  \begin{tabular}{clc}
    \hline
  \# \textbf{Label} & \textbf{Describe} & \textbf{Example}\\
    \hline
    0.0 & Other & haha cái này vui nha \\
    1.0 & Discrimination& bắc kỳ toàn lũ kh*n n*n \\
    2.0 & Supportive & Nam Bắc một nhà\\
    \hline
    \label{table example data_V}
  \end{tabular}
\end{table}
Label 0.0 (Other): This category includes comments that may be insignificant remarks or comments that cannot be classified as 1.0 or 2.0.

Label 1.0 (Discriminatory): This category includes comments containing offensive language directed at a specific group of people in a particular location or region.

Label 2.0 (Supportive): This category includes comments that oppose regional discrimination or express respect for the culture and people of different regions.

\subsubsection{Building dataset}
\begin{figure}[h]
    \centering
    \includegraphics[width=1\linewidth]{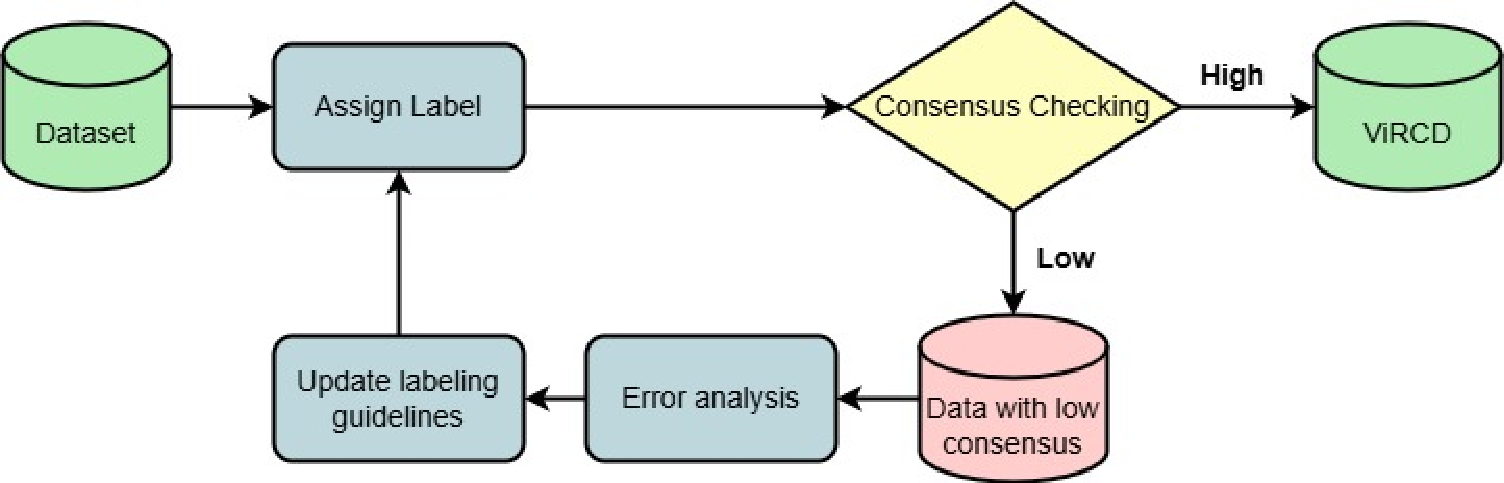}
    \captionsetup{font=footnotesize}
    \caption{Data Labeling Process }
    \label{fig: data labeling process}
\end{figure}

We labeled the dataset using the process shown Fig \ref{fig: data labeling process}. First, we collected data from user comments on social media platforms. After gathering the data, we proceeded with the labeling process. Comments with low consensus among team members were relabeled. Additionally, we conducted error analysis on low-consensus cases and updated our labeling guidelines to enhance the dataset's quality. 

\renewcommand{\figurename}{Figure}

\subsubsection{Example}
Two example tables (Table \ref{table: 3} and Table \ref{table 4}) written in English and Vietnamese:
\renewcommand{\tablename}{Table}
\begin{table}[H]
\centering
\caption{Some examples in dataset}
  \renewcommand{\arraystretch}{1.5}
  \begin{tabular}{clc}
    \hline
    \# & \textbf{Comment} & \textbf{Label} \\
    \hline
    1 & F*ck Southerners  & 1.0 \\
    % 2 & Parky is very irritating & 1.0 \\
    2 & North, Central, South - one family & 2.0 \\
    3 & the guy in the white shirt speaks so well & 0.0 \\
    4 &Thanh Hóa people live deceitfully, cunningly... & 1.0 \\
    \hline
  \end{tabular}
  \label{table: 3}
\end{table}
\renewcommand{\tablename}{Table}
\begin{table}[H]
\centering
\caption{Some examples in the dataset are presented in Vietnamese}
\renewcommand{\arraystretch}{1.5}
\begin{tabular}{clc}
  \hline
  \# & \textbf{Comment} & \textbf{Label} \\
  \hline
  1 & đ*t\_mẹ bọn nam\_kỳ & 1.0 \\
  % 2 & Parky cayyy & 1.0 \\
  2 & Bắc Trung Nam một nhà & 2.0 \\
  3 & anh áo trắng nói hay quá & 0.0 \\
  4 & thanh hóa sống bạc lừa\_lọc ranh\_mãnh... & 1.0 \\
  \hline
\end{tabular}
\label{table 4}
\end{table}

\subsection{Data preprocessing}
\begin{figure}[H]
    \centering
    \includegraphics[width=1\linewidth]{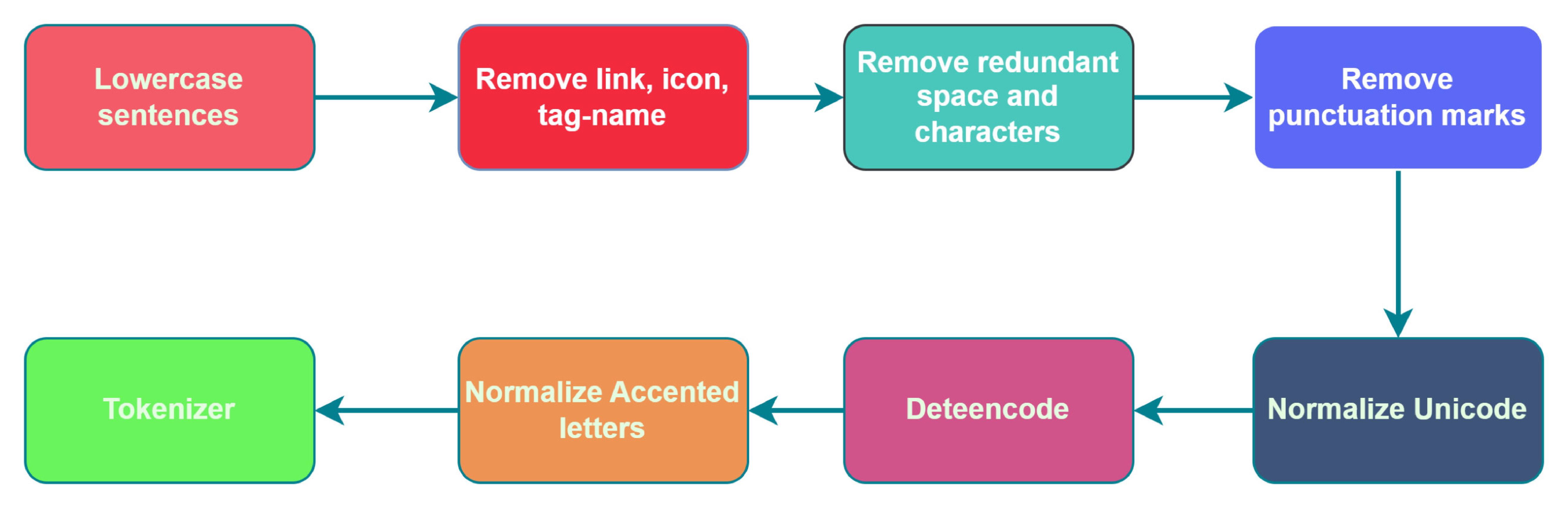}
    \captionsetup{font=footnotesize}
    \caption{Preprocessing Steps}
    \label{fig:preprocessing_3}
\end{figure}

An important step in NLP is data preprocessing. Data preprocessing is the process of preparing raw text data for further analysis. This process helps remove errors and noise from the data while also converting it to a format that is easier for NLP models to understand. We process raw data through each stage in Fig \ref{fig:preprocessing_3}.

\renewcommand{\figurename}{Figure}

- \textbf{Lowercase Sentences}: We lowercase all the data points' characters in the dataset. We do this to avoid Python seeing two exact words as separate because of their forms.

- \textbf{Delete link, icon, tag name}: In many posts, We have seen many comments containing tag names, icons, and links to other websites. So we decide to remove all of these from our data.

- \textbf{Delete Redundant Space/Character}: Users on social media unwittingly or wittingly type multiple spaces and repeat characters within their comments. As a result, we decide to remove redundant spaces/characters.

- \textbf{Delete punctuation marks}: In the environment of social networks, punctuation marks, and special characters are used extensively to give a more lively and emotional meaning to the reader. 

- \textbf{Normalize Unicode}: We see a lot of Vietnamese words in the dataset that are the same, but Python sees them as separate because of its Unicode. There are some popular Unicode in comments like UTF-8, UTF-16BE, UTF-16LE, UTF-32…. So we decide to normalize all to UTF-8.

- \textbf{De-teencode and decode abbreviate}: there are many teen codes and abbreviations in comments so we decided to detect and decode these to correct word.

Example:
\begin{itemize}[\leftmargin=5.0em]
  \item đc $\rightarrow$ được
  \item k $\rightarrow$ không
\end{itemize}

- \textbf{Normalize Accented Letters}: the Vietnamese language on social media has many different ways of typing accents for example "quà" and "qùa", "hóa" and "hoá". These are the same word with the same meaning but when put into Python, it will recognize these are different letters. So it is essential to normalize these words into one form. 

- \textbf{Word Tokenizer}: The input sentence is split into words or meaningful word phrases. To utilize this, we use the Word Segmenter of VnCoreNLP to our dataset for training the PhoBert model.

Example:
\begin{itemize}[\leftmargin=5.0em]
  \item Việt Nam $\rightarrow$ Việt\_Nam
  \item Phân biệt $\rightarrow$ phân\_biệt
\end{itemize}

% \renewcommand{\figurename}{Figure}
% \begin{figure}[H]
%     \centering
%     \includegraphics[width=1.0\linewidth]{Figures/3_Data/ploai.jpg}
%     \caption{Example of preprocessing steps}
%     \label{fig:ploai}
% \end{figure}
% \subsection{Dataset Overview}
% \label{subsec: Dataset Overview}

\begin{figure}[h!]
    \centering
    \includegraphics[width=0.7\linewidth]{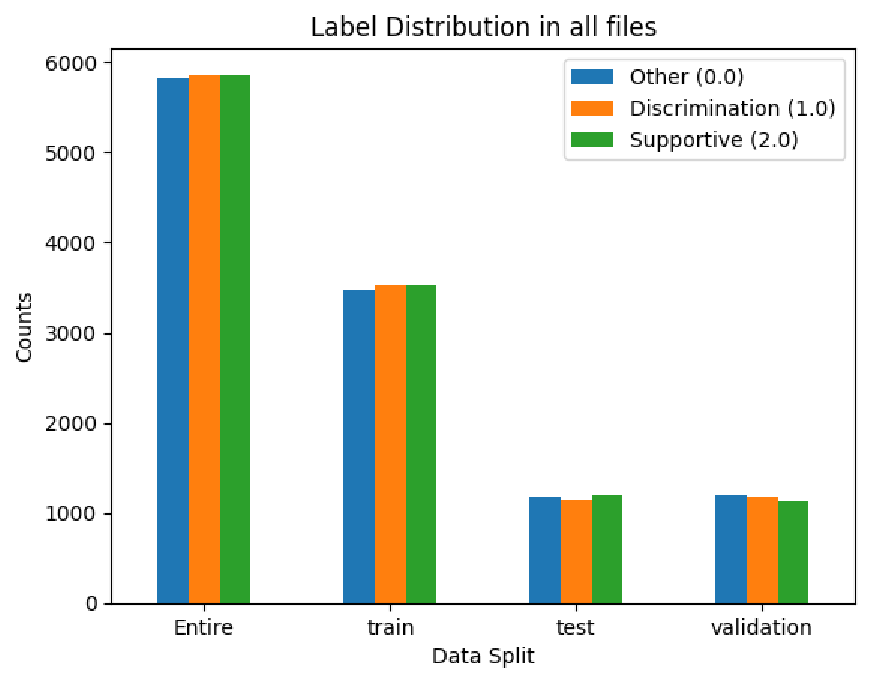}
    \captionsetup{font=footnotesize}
    \caption{Label Distribution in all Files}
    \label{fig:label distribution}
\end{figure}

We balanced three labels in the dataset mainly to improve model performance and avoid biased predictions after balances We then divided the data into three sets of train validation and test to train the language models for solving the problem in Fig \ref{fig:label distribution} \\
% The number of each label in the dataset:

\renewcommand{\figurename}{Figure}

\begin{figure}[h!]
    \centering
    \includegraphics[width=0.9\linewidth]{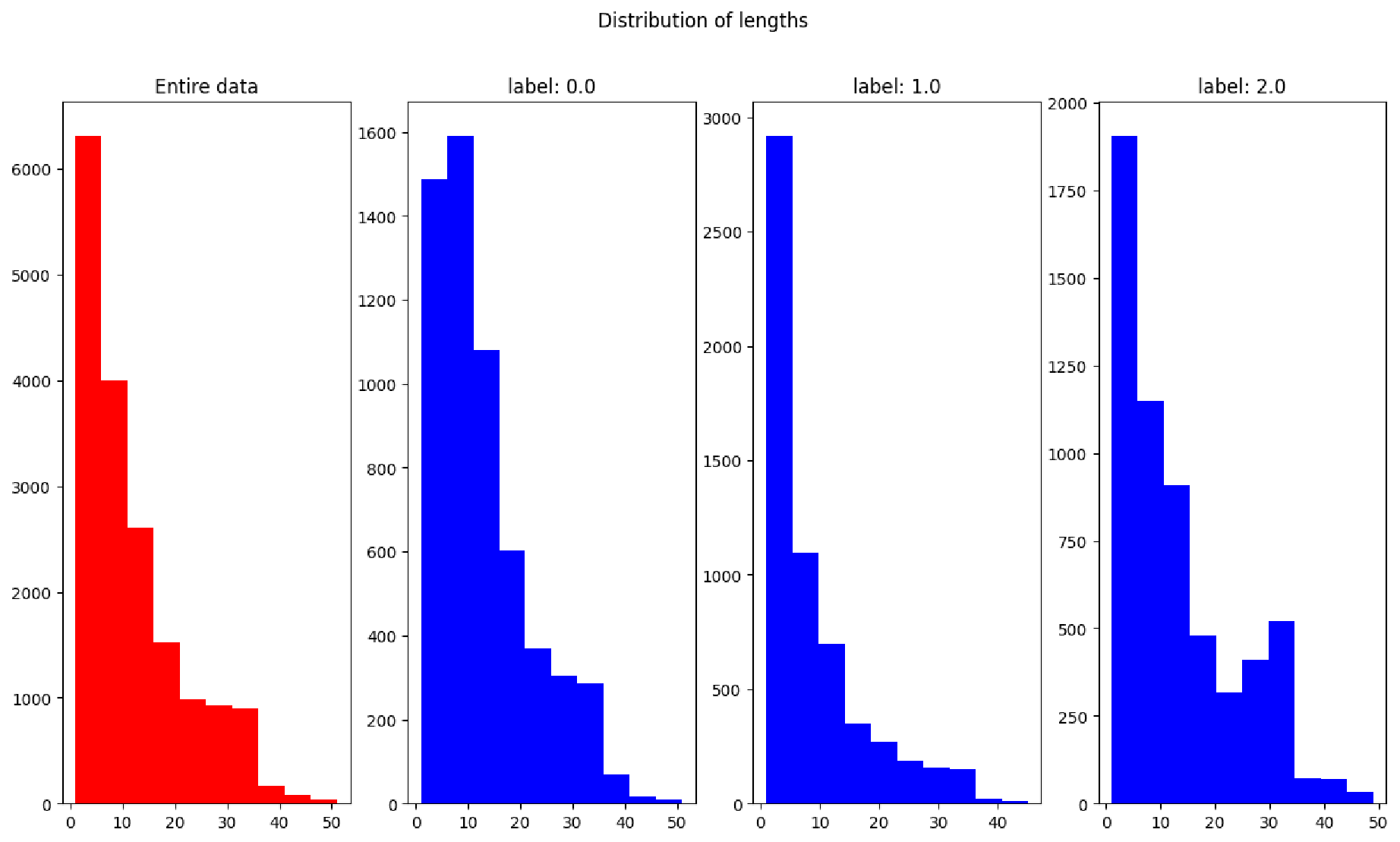}
    \caption{Comment Length Distribution}
    \captionsetup{font=footnotesize}
    \label{fig:word count}
\end{figure}

We first analyzed sentence lengths to find differences among labels but found the average word count was similar: 12,388 for label 0.0, 8,599 for 1.0, and 10,384 for 2.0. With no significant discrepancies (Fig \ref{fig:word count}), So we explored other characteristics, focusing on the most frequent words in each label using word clouds.
% We first analyzed sentence lengths to find differences among labels but found the average word count was similar: 12,388 for label 0.0, 8,599 for 1.0, and 10,384 for 2.0. With no significant discrepancies (Fig \ref{fig:word count}), So we explored other characteristics, focusing on the most frequent words in each label using word clouds. Fig \ref{fig: word cloud} contain subfigures (a, b, c, d) that illustrate these insights.

\renewcommand{\figurename}{Figure}

% \begin{figure}[h!]
%     \centering
%     \begin{tabular}{cc}
%         \begin{subfigure}{0.2\textwidth}
%     \includegraphics[width=\textwidth]{Figures/3_Data/cmtpbvm.jpg}
%     \caption{in Dataset}
%             \end{subfigure}&
%             \begin{subfigure}{0.2\textwidth}
%                  \includegraphics[width=\textwidth]{Figures/3_Data/label00.jpg}
%                  \caption{in label 0.0}
%             \end{subfigure}\\
%             \begin{subfigure}{0.2\textwidth}
%               \includegraphics[width=\textwidth]{Figures/3_Data/label10.jpg}
%               \caption{in label 1.0}
%             \end{subfigure}&
%             \begin{subfigure}{0.2\textwidth}
%         \includegraphics[width=\textwidth]{Figures/3_Data/label20.jpg}
%         \caption{in label 2.0}
%             \end{subfigure}
%         \end{tabular}
%     \label{fig: word cloud}
%     \caption{ The Most Frequently Occurring Words  }
% \end{figure}
During our analysis, we observed that the length distributions of comments do not significantly differ across labels. Still, the opposite is true for the popularity of words within sentences. Regional discrimination label typically often include terms like parky, namky, and California. On the other hand, supportive comments tend to use common phrases such as Bắc Trung (North-Central), Nam Bắc (South-North), Trung Nam (South-Central), and some words related to the unity of a country like tổ quốc (Motherland), một nhà (As one), đồng bào (Compatriots)'. Finally, comments labeled as "other" generally often utilize common words like không (no), and bọn (Gang).

\section{Model}
\subsection{Machine-learning model}
\subsubsection{Random Forest}

Random Forest is an ensemble learning method that builds multiple decision trees and outputs the mode of the trees' predictions. It's known for its robustness, ability to handle large, high-dimensional datasets, and resistance to overfitting. In comment classification, Random Forest can capture complex text patterns, making it highly accurate.

\subsubsection{Multinomial Logistic Regression}

Multinomial Logistic Regression extends Logistic Regression to handle multiple classes. Unlike binary logistic regression, it models the probability distribution across several outcomes, making it suitable for multi-class problems.
\subsubsection{Multinomial Naive Bayes}

Multinomial Naive Bayes is tailored for multinomially distributed data, ideal for text classification. It assumes features (words) are generated from a multinomial distribution, fitting naturally with word counts in documents.
\subsection{Transfer Learning Model}

% Transformers is a deep learning model introduced in 2017, used mainly in the fields of natural language processing (NLP) and computer vision (CV). Unlike RNN, Transformers do not require sequential data to be processed in order. This means transformer has training parallelization capability that allows training on larger data. It is increasingly the choice for NLP problems to replace RNN models such as Long-Short Term Memory (LSTM). And BERT is a model built upon the Transformer architecture and it has achieved impressive results on many natural language tasks. Taking advantage of the superior capabilities of the transformer architecture, we apply deep learning models with the transformer architecture to perform classification for this problem 

% \renewcommand{\figurename}{Figure}
% \begin{figure}[h!]
%     \centering
%     \includegraphics[width=0.6\linewidth]{Figures/4_Model/deeplearning.jpg}
%     \caption{The Transformer model architecture}
%     \label{fig:deeplearning}
% \end{figure}

\subsubsection{PhoBERT}

 The transfer learning model has attracted increasing attention from NLP researchers around the world for its outstanding performance. One of the SOTA language models BERT, which stands for Bidirectional Encoder representations from transformers, is published by Devlin et al\cite{devlin2018bert}. For Vietnamese, the SOTA method was first released and called PhoBERT by Nguyen et al\cite{nguyen2020phobert}. for solving Vietnamese NLP problems. PhoBERT is a pre-trained model, which has the same idea as RoBERTa, a replication study of BERT was released by Liu et al\cite{liu2019roberta}, and there are modifications to suit Vietnamese.

% \bibliographystyle{plain}
% \bibliography{references}
% PhoBERT is a pre-training with monolingual language training, which means it is only trained specifically for Vietnamese, developed by the research team at VinAI Research. PhoBERT has the same approach as RoBERTa - an improvement of BERT. PhoBERT is trained on about 20GB of data (including about 1GB Vietnamese Wikipedia corpus and 19GB from Vietnamese news corpus). This is a fairly decent amount of data, allowing the model to understand and work effectively with the diverse contexts and meanings of Vietnamese today.
\subsubsection{XLM-RoBERTa}
  A large multi-lingual language model, developed by Facebook AI Research \cite{XML_Roberta}. This model is an extended and improved version of the RoBERTa model. It has trained on 2.5TB of filtered data from Common Crawl (including Vietnamese and other languages worldwide).
\subsubsection{BERT-Base Multilingual Cased}
With BERT-base \cite{DBLP:journals/corr/abs-1810-04805}, there are two multilingual models currently available (Cased/Uncased). The model is pre-trained on the top 104 languages with the largest Wikipedia using a masked language modeling (MLM) objective in a self-supervised fashion.
\section{Experiments and Results}
\label{sec: EXPERIMENTS AND RESULTS}
\subsection{Experiments proposed system}

\begin{figure}[h!]
    \centering
    \includegraphics[width=1.0\linewidth]{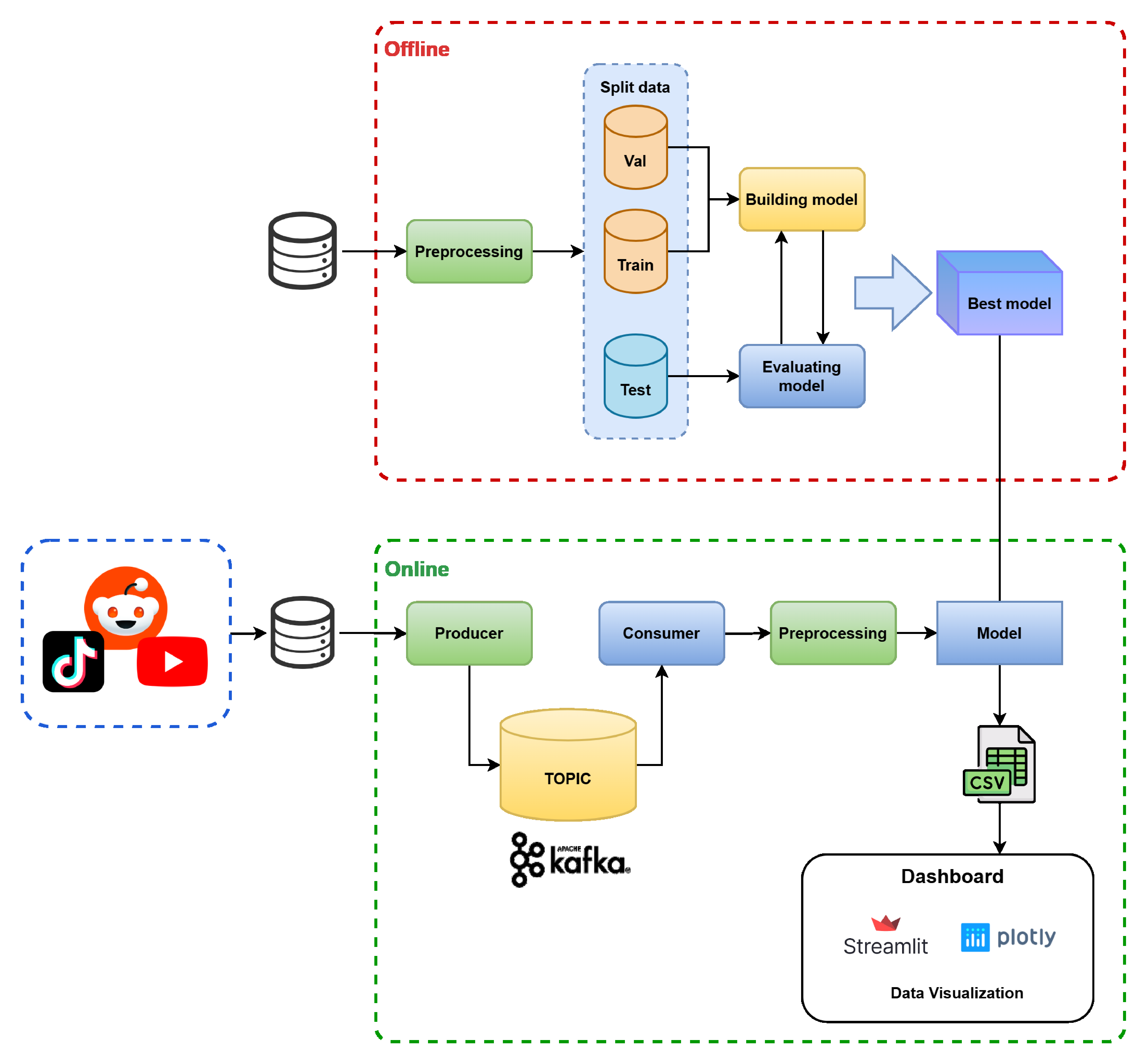}
    \captionsetup{font=footnotesize}
    \caption{The architecture of the proposed system}
    \label{fig: proposed system}
\end{figure}

{To address this issue, we have proposed a system as shown in Figure \ref{fig: proposed system}}. 

{The system consists of two main components. In the offline system. First, the dataset will be preprocessed before being divided into three parts: training, testing, and validation, to perform model building and performance evaluation. Next, we select the model with the highest performance to be deployed in the online system for streaming data processing. In the online system, after crawling data from Facebook and TikTok, data will go through the processing steps of Kafka and Spark and be preprocessed simultaneously.  The best model will get data to predict the preprocessed data, The result will be stored in a CSV file and visualized on the dashboard by Streamlit and Matplotlib }

\subsection{Experiment settings}

As we analyzed in the Data section, the labels appear to be more discriminative based on the frequency of word occurrences in a comment sentence rather than the number of words in those comments. Therefore, we utilize CountVectorizer from the scikit-learn library for our machine-learning models to detect key phrases in the comments and extract their features. 

We also use GridSearchCV for the configuration of machine-learning models in this project:

\begin{itemize}
    \item \textbf{Random Forest:} We implement a Random Forest Classifier model with n\_estimators = 400, criterion = `entropy', and random\_state = 42.
    \item \textbf{Multinomial Logistic Regression:} This model is run with multi\_class = `multinomial', random\_state = 42, solver = `lbfgs'. 
    \item \textbf{Multinomial Naive Bayes:} We use MutinomialNB Classifier with `alpha' = 1.0, it is the default value of this model. In addition, random\_state = 42.
\end{itemize}

In addition, transfer learning models were employed to address this classification problem. Due to hardware constraints, all experiments utilized the base versions of the models. Specifically, \textbf{PhoBERT}, \textbf{XLM-RoBERTa}, and \textbf{BERT-base-multilingual-cased} were configured with five epochs, a maximum sequence length of 512 to capture the full content of posts and comments, a batch size of 8, and a learning rate of 5e-06.

\subsection{Evaluate metrics}

In this problem, we focus on two main evaluation metrics: Accuracy and F1-macro. However, during the evaluation process, we place more emphasis on the F1-macro metric. The focus on F1-macro is due to the class imbalance within the dataset. When a dataset is imbalanced, evaluating solely based on Accuracy can lead to misunderstandings. Accuracy can be high when the model focuses only on predicting the majority class and disregards the minority class. This does not accurately reflect the model’s classification ability.  

\begin{center}
    \large
     Accuracy = $\frac{\text{Number of correct predictions}}{\text{Total number of predictions}}$ \\
    \vspace{19pt}
    F1-macro = $\frac{2 \times \text{Recall} \times \text{Precision}}{\text{Recall} + \text{Precision}}$ \\
\end{center}

By emphasizing F1-macro, we ensure that the model performs well across all classes, including the minority classes. This implies that the model has the ability to classify accurately and consistently across different instances, ensuring fairness and reliability in the evaluation process 

\subsection{Experiment results}

\begin{table}[h]
    \centering
    \caption{Evaluation results on the Vietnamese Regional Discrimination Comments datasets}
    \renewcommand{\arraystretch}{1.5}
    \begin{tabular}{lcc}\hline
         \textbf{Models}&  \textbf{Accuracy}&  \textbf{F1-score}\\
         \hline
         \textbf{Random Forest}& \textbf{0.9712 } &  \textbf{0.9700}\\
         \textbf{Multinomial Logistic Regression}& 0.9108 & 0.9100\\
         \textbf{Multinomial Naive Bayes}& 0.8400 & 0.8300\\ 
         \hline
         \textbf{XML-RoBERTa}& 0.8979 & 0.8902\\
         \textbf{PhoBert}&  0.9206&  0.9187\\  
         \textbf{Bert-base-multilingual-cased}& 0.9174 & 0.9154  \\ 
         \hline
    \end{tabular}
    \label{table: 5}
\end{table}

\subsection{Results analysis}

\begin{figure}[h!]
    \centering
    \includegraphics[width=0.8\linewidth]{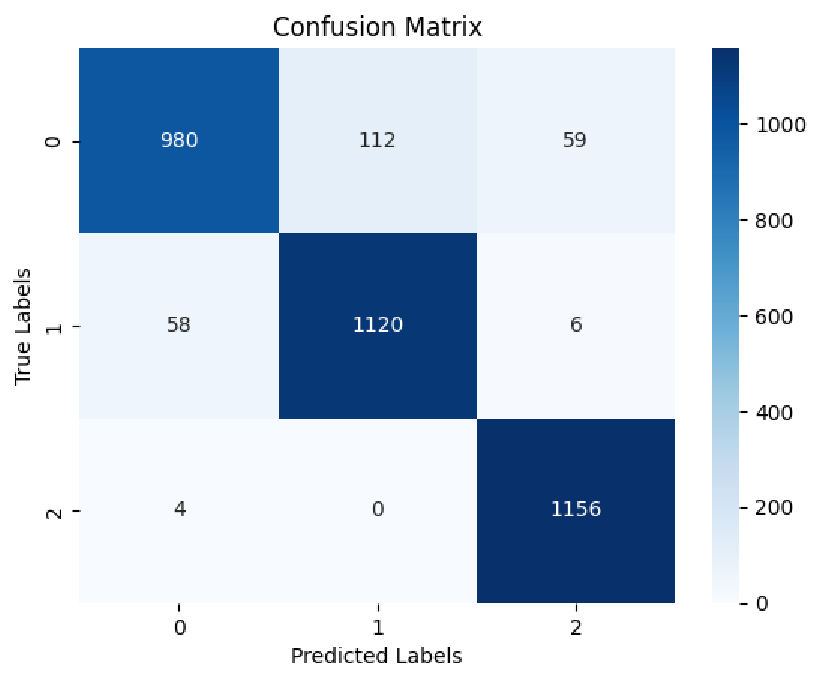}
    \caption{Confusion matrix}
    \captionsetup{font=footnotesize}
    \label{fig: confusion matrix}
\end{figure}

\begin{table}[h]
\centering
\caption{Some examples of classification error}
\renewcommand{\arraystretch}{1.5}
\begin{tabular}{lcc} 
    \hline 
    \textbf{Content} & \textbf{True label} & \textbf{Predict}\\
    \hline
     người phân biệt là người miền nam & 1.0 & 0.0\\
     parky tht ra là từ tiếg anh thật mà &  0.0 & 1.0\\
     park ji sung&  0.0& 1.0\\
     bawts kyf ca&  1.0& 0.0\\
    bắcki	& 1.0&0.0\\
    sao paky nói zây& 1.0&0.0\\
    \hline
\end{tabular}
\label{tab:my_label}
\end{table}

After obtaining the results in Table \ref{table: 5} We found that Random Forest with the ensemble learning method performs exceptionally well on the dataset due to its ability to partition the data space into smaller segments, thereby easily detecting differences based on the specific features of each comment. We analyzed the errors of the highest-performing model. Fig \ref{fig: confusion matrix} shows the confusion matrix of the model on the test set. We can observe that the two labels most frequently confused with each other are "Discrimination" and "Other". Additionally, the label  "Supportive" is well classified by the model.  

Sentences contain discriminate words but don't have means to discriminate like the word "parky" in "parky that ra là từ tiếng anh thật mà" or likely discrimination words but not and such as park, party... may confuse the model 

On the contrary, certain sentences devoid of explicit regional discriminatory or negative language often harbor concealed stereotypes about individuals from diverse regions. Such instances pose challenges to the model's classification capabilities.

Some misspellings in sentences also cause errors for the model like "bắcki", "iwr mỹ cũng gọi nam cali bắc cai cis sao đâu" or Keyboard input methods for creating diacritics for Vietnamese words and phrases as defined by UniKey software, which is not encoded in comments like "bawt kyf ca" instead of "bắc kỳ ca".
\section{Streaming}
 \subsection{Data Streaming}
{Data streaming is the process of continuously transmitting data without interruptions. Instead of processing data once, like traditional methods, data streaming processes data as soon as it is received and continues while the data is still being transmitted. This allows companies and organizations to quickly receive processing results based on the latest data.
Currently, there are many popular data streaming systems, such as Apache Kafka, Apache Spark Streaming, Amazon Kinesis, etc., actively supporting real-time data processing. In this report, the research team will apply Apache Kafka to the group's streaming system.}

\subsection{Streaming Architecture}
\begin{figure*}[h!]
    \centering
    \includegraphics[width=0.7\linewidth]{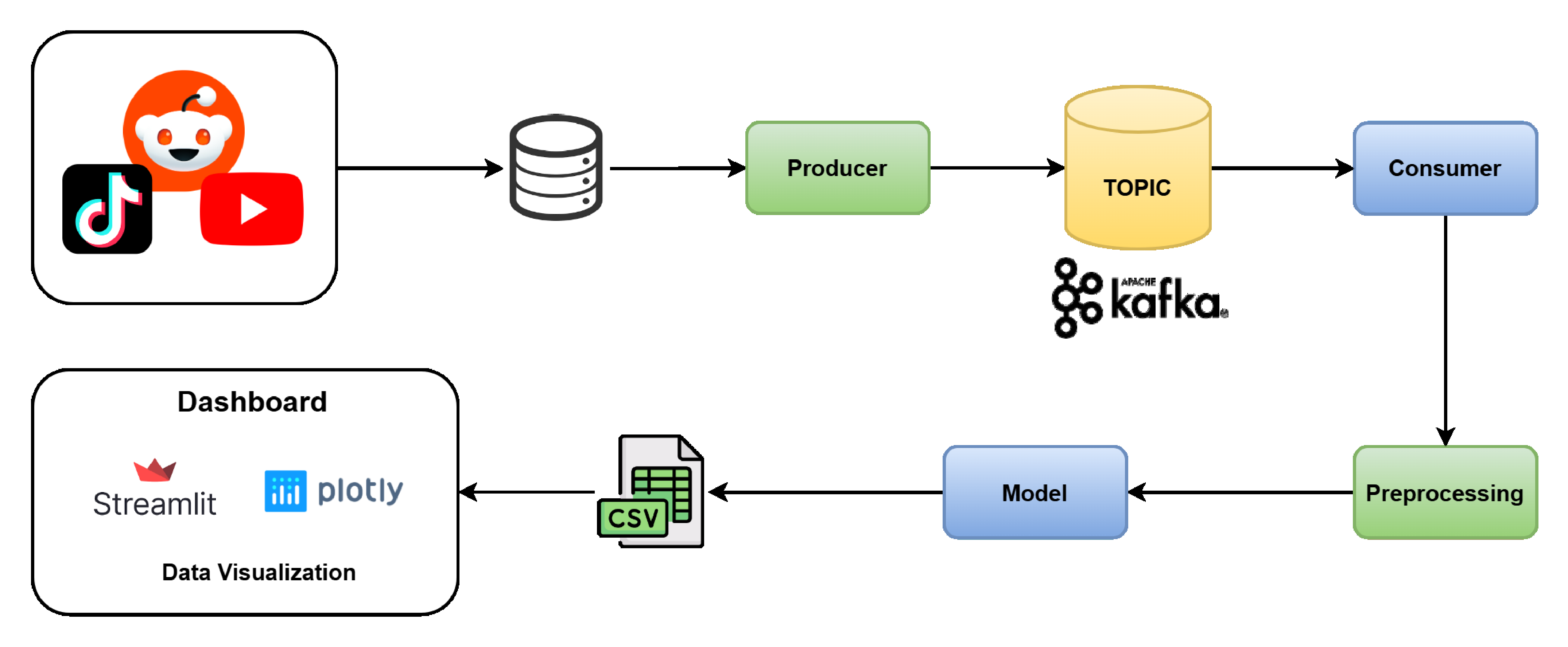}
    \caption{The Architecture of Streaming System}
    % \captionsetup{font=footnotesize}
    \label{fig: streaming flow}
\end{figure*}

{The group's processing flow is divided into two main parts, as described in the Figure \ref{fig: streaming flow} }

\textbf{Part 1: Data Collection from Facebook and TikTok and Sending It Through Kafka Producer:}

{Firstly, we collect comments from posts on social media platforms. Additionally, we also built a Kafka producer to send the UTF-8-encoded comments through the created topic.}

\textbf{Part 2: Preprocessing and Classification of Comments Using the Random Forest Classifier Model:}

{After conducting experiments and comparing different models in Part \ref{sec: EXPERIMENTS AND RESULTS}, we have chosen the Random Forest Classifier Model as the classification model used in our system. Next, we create a Kafka Consumer with the created topic, decode the received messages using UTF-8, and feed them into the preprocessing process. Then, we perform data vectorization to match the model input. After the data encoding process, the data will be stored as CSV files, and these data will be provided for analysis and statistics. }

We have also built an interface for the data classified by the best model to visualize the results and analyze them with tables and charts: 

% \begin{figure}[h!]
%     \centering
%     \includegraphics[width=0.75\linewidth]{Figures/6_Streaming/visualize(1).png}
%     \caption{Data Table}
%     \label{fig:enter-label}
% \end{figure}

% \begin{figure}[h!]
%         \centering
%         \includegraphics[width=0.75\linewidth]{Figures/6_Streaming/visualize(3).png}
%         \caption{Top 20 words in Label Discrimination}
%         \label{fig:enter-label}
%     \end{figure}
% \begin{figure}[h!]
%         \centering
%         \includegraphics[width=0.75\linewidth]{Figures/6_Streaming/visualize(2).png}
%         \caption{Result's Pie Chart}
%         \label{fig:enter-label}
%     \end{figure}

% \begin{figure}[h!]
%     \centering
%     \includegraphics[width=0.8\linewidth]{Figures/6_Streaming/image.png}
%     \caption{Users can input File or String for the Model to predict Regional Discrimination}
%     \label{fig:enter-label}
% \end{figure}
% Users can use the dashboard to monitor the crawled and predicted data in the system. They can also import a string of sentences or a CSV file with column names set according to the software's instructions to classify and visualize the data 
\section{Conclusion and Future work }

In this paper, we implemented and compared text classification models to assess their performance. We introduced the novel dataset ViRDC which comprises comments from social media platforms for Regional Discrimination Detection. We also build a system that can process real-time data from social media networks to ensure the scalability and responsiveness of the system with the Apache Spark framework to handle the increasing data inputs. 

However, our ambitions extend beyond mere model performance. We aim to revolutionize real-time comment classification on social media networks using Apache Kafka, with a focus on scalability and efficiency. Our future plans include leveraging advanced natural language processing models to detect complex discrimination types, such as regional, gender, disability, national, and racial disparities. Refining our labeling process will be crucial for enhancing dataset quality and performance. Additionally, we intend to integrate various frameworks, including deep learning and transfer learning models, with fewer hardware limitations to handle larger and more intricate data, thereby improving classification capabilities for complex problems.

In the long run, we're looking to create a system interface that's easy to use and can blend effortlessly with the platforms people already use. This approach will make our system both accessible and practical. By following these future plans, we hope to make big improvements in how comments are classified and the wider effects this can have.

\bibliographystyle{IEEEtran}
% \bibliography{sn-bibliography} 
% \bibliographystyle{unsrt}
\bibliography{references.bib}

\end{document}